%% file: main.tex
\documentclass{article}
\usepackage{spconf,amsmath,graphicx}
\usepackage{tikz}
\usepackage{amssymb}
\usepackage{amsthm}
\usepackage{lipsum}
\usepackage{float}
\usepackage{booktabs}
\usepackage{adjustbox}
\usepackage{hhline}
\usepackage{subfig}
\usepackage{bm}
\usepackage{tabularx}
\newcolumntype{Y}{>{\centering\arraybackslash}X}
\usepackage[font=footnotesize,labelfont=bf]{caption}
\usepackage{multirow}
\usepackage{siunitx}

\usepackage{float}
\usepackage{hhline}
\restylefloat{table}
\usepackage{arydshln}
\usepackage{makecell}
\usepackage{dsfont}
\usepackage{upgreek}

\usepackage{comment}
\usepackage[section]{placeins}

\usepackage[numbers,sort&compress]{natbib}
\usepackage{enumitem}
\usepackage{hyperref}
\setlist[description]{labelindent=1cm,leftmargin=3cm,style=multiline}

\usepackage{letltxmacro}
\usetikzlibrary{positioning}
\tikzset{main node/.style={circle,fill=blue!20,draw,minimum size=1cm,inner sep=0pt},
            }
            
\newcommand{\stargraph}[3]{\begin{tikzpicture}
    \node[circle,fill=red] at (360:0mm) (center) {};
    \foreach \n in {1,...,#1}{
        \node[circle,fill=black] at ({\n*360/#1}:#2cm) (n\n) {};
        \node at (0,-#2*1.5) {#3}; 
    }
    \node[circle,draw=red, scale=1.5] at ({0}:#2cm) (max) {};
    \draw (center) edge[red, ultra thick] node[left] {} (max);
\end{tikzpicture}}

\newcommand{\fullygraph}[3]{\begin{tikzpicture}
    \node[circle,fill=red] at (360:0mm) (center) {};
    \foreach \n in {1,...,#1}{
        \node[circle,fill=black] at ({\n*360/#1}:#2cm) (n\n) {};
        \draw (center)--(n\n);
        \foreach \m in {1,...,\n}{
            \draw (n\n)--(n\m);
        };
    \node[circle,draw=red, scale=1.5] at ({0}:#2cm) (max) {};
        \node at (0,-#2*1.5) {#3}; 
    }
\end{tikzpicture}}

\newcommand{\fullyredstar}[3]{\begin{tikzpicture}
    \node[circle,fill=red] at (360:0mm) (center) {};
    \foreach \n in {1,...,#1}{
        \node[circle,fill=black] at ({\n*360/#1}:#2cm) (n\n) {};
        \foreach \m in {1,...,\n}{
            \draw (n\n)--(n\m);
        };
    \node[circle,draw=red, scale=1.5] at ({0}:#2cm) (max) {};
    \draw (center) edge[red, ultra thick] node[left] {} (max);
        \node at (0,-#2*1.5) {#3}; 
    }
\end{tikzpicture}}

\ninept
\makeatletter
\let\oldr@@t\r@@t
\def\r@@t#1#2{%
	\setbox0=\hbox{$\oldr@@t#1{#2\,}$}\dimen0=\ht0
	\advance\dimen0-0.2\ht0
	\setbox2=\hbox{\vrule height\ht0 depth -\dimen0}%
	{\box0\lower0.4pt\box2}}
\LetLtxMacro{\oldsqrt}{\sqrt}
\renewcommand*{\sqrt}[2][\ ]{\oldsqrt[#1]{#2}}
\makeatother

\makeatletter
\newcommand{\thickhline}{%
	\noalign {\ifnum 0=`}\fi \hrule height 1pt
	\futurelet \reserved@a \@xhline
}
\newcolumntype{"}{@{\hskip\tabcolsep\vrule width 1pt\hskip\tabcolsep}}
\makeatother

\makeatletter
\newcommand*{\defeq}{\mathrel{\rlap{%
			\raisebox{0.3ex}{$\m@th\cdot$}}%
		\raisebox{-0.3ex}{$\m@th\cdot$}}%
	=}
\makeatother

\newcommand{\smpar}[1]{\vspace{0.7em}\noindent\textbf{#1:}}
\newcommand{\vect}[1]{\boldsymbol{#1}}

\def\R{{\mathbb{R}}}

\def\K{{\mathbf{K}}}

\def\softmax{{\boldsymbol{\upsigma}}}

\title{On the Structures of Representation for the Robustness of Semantic Segmentation to Input Corruption}
\name{Charles Lehman, Dogancan Temel, and Ghassan AlRegib}
\address{OLIVES at the Center for Signal and Information Processing,\\ School of Electrical and Computer Engineering,\\ Georgia Institute of Technology, Atlanta, GA, 30332-0250 USA\\ \{charlie.k.lehman,cantemel,alregib\}@gatech.edu}
\begin{document}

\input{_4_sections/__arxiv_cover.tex}

\maketitle
\begin{abstract}
\input{_4_sections/_0_abstract.tex}

\end{abstract}
\input{_4_sections/_1_introduction.tex}
\input{_4_sections/_2_background.tex}

\input{_4_sections/_3_method.tex}
\input{_4_sections/_4_experiment.tex}
\input{_4_sections/_5_conclusion.tex}
\bibliographystyle{IEEEbib}
\bibliography{main}
\end{document}

%% file: _4_sections/__arxiv_cover.tex
\onecolumn 

\begin{description}

\item[\textbf{Citation}]{C. Lehman, D. Temel and G. AlRegib, "On the Structures of Representation for the Robustness of Semantic Segmentation to Input Corruption," IEEE International Conference on Image Processing (ICIP), Abu Dhabi, United Arab Emirates, Oct. 2020.
} \\



\item[\textbf{Code}]{\url{https://github.com/olivesgatech/segmentation_corruption}} \\

\item[\textbf{Bib}] {
@INPROCEEDINGS\{Lehman2020,\\ 
author=\{C. Lehman and D. Temel and G. AIRegib\},\\ 
booktitle=\{IEEE International Conference on Image Processing (ICIP)\},\\ 
title=\{On the Structures of Representation for the Robustness of Semantic Segmentation to Input Corruption\},\\ 
year=\{2020\},\}\\
} \\

\item[\textbf{Copyright}]{\textcopyright 2020 IEEE. Personal use of this material is permitted. Permission from IEEE must be obtained for all other uses, in any current or future media, including reprinting/republishing this material for advertising or promotional purposes,
creating new collective works, for resale or redistribution to servers or lists, or reuse of any copyrighted component
of this work in other works. } \\

\item[\textbf{Contact}]{\href{mailto:alregib@gatech.edu}{alregib@gatech.edu}~~~~~~~\url{https://ghassanalregib.com/} \\ \href{mailto:charlie.k.lehman@gmail.com
}{charlie.k.lehman@gmail.com~~~~~~~\url{https://charlielehman.github.io/}
} \\ \href{mailto:dcantemel@gmail.com}{dcantemel@gmail.com}~~~~~~~\url{http://cantemel.com/}}
\end{description} 

\thispagestyle{empty}
\newpage
\clearpage

\twocolumn

%% file: _4_sections/_0_abstract.tex
Semantic segmentation is a scene understanding task  at the heart of safety-critical applications where robustness to corrupted inputs is essential.  
Implicit Background Estimation (IBE) has demonstrated to be a promising technique to improve the robustness to out-of-distribution inputs for semantic segmentation models for little to no cost.
In this paper, we provide analysis comparing the structures learned as a result of optimization objectives that use Softmax, IBE, and Sigmoid in order to improve understanding their relationship to robustness.
As a result of this analysis, we propose combining Sigmoid with IBE (SCrIBE) to improve robustness.
Finally, we demonstrate that SCrIBE exhibits superior segmentation performance aggregated across all corruptions and severity levels with a mIOU of 42.1 compared to both IBE 40.3 and the Softmax Baseline 37.5.

%% file: _4_sections/_1_introduction.tex
\section{Introduction}
\vspace{-0.5em}
\label{sec:intro}
In the past two years alone, there has been explosive growth in automated applications built upon advances made in deep learning \cite{Zion2019}.  
For vision systems alone, deep learning has paved the way to a host of new products and services in safety-critical applications from autonomous vehicles to medical diagnosis to surveillance \cite{hatcher2018survey}.  
Largely attributed to the increase in accessibility of open-source software and computing power \cite{pytorch2019, tensorflow2015-whitepaper}, the barriers between lab-born innovations and market-ready products are lower than ever before. Though this results in the ability to bring deep vision systems to market quickly, it may be premature for safety-critical applications \cite{ntsbHWY18MH010, KamannBenchmarkingRobustnessSemantic2019}.
For applications where the failure of the vision system can result in severe consequences---such as, damage or harm---it is necessary that they are robust.
In essence, robustness ensures that a system can prevent or minimize the impact of failures. Despite being of limited use in safety-critical applications, much of the work toward robustness of deep vision is centered around image classification \cite{hendrycks17baseline, liang2018enhancing, lee2018simple, hendrycks2018oe, guo2017calibration, lee2018training, hendrycks2018benchmarking, Carlinievaluatingrobustnessneural2017, papernot2016limitations, GoodfellowExplainingharnessingadversarial2014,Temel2018_CUREOR, temel2019traffic, GeirhosGeneralisationhumansdeepa, Temel2017_NIPSW,Temel2018_SPM,Temel2019_ICIP}.

To be more relevant to real-world need, we study semantic segmentation, which is at the heart of safety-critical decision systems across a broad spectrum of applications due to simultaneously performing localization and classification.
Despite the dizzying pace of advancements for semantic segmentation, contributions have largely only been toward improving task performance on increasingly challenging datasets \cite{EveringhamPascalVisualObject2015, LinMicrosoftCOCOCommon2015, Cordts2016Cityscapes, ZhouSceneparsingade20k2017} or reducing the resource-footprint to capability ratio \cite{SiamRTSegRealTimeSemantic2018, MehtaESPNetEfficientSpatial2018, HowardMobileNetsEfficientConvolutional2017, PaszkeENetDeepNeural2016}.
Though still limited, there have been recent contributions to semantic segmentation for assessing and improving robustness to adversarial \cite{ZhouAutomatedEvaluationSemantic2019}, out-of-distribution \cite{lehman2019implicit}, and corrupted inputs \cite{KamannBenchmarkingRobustnessSemantic2019}.
Though we expand upon the techniques from \cite{lehman2019implicit} the most similar work to the presented work in this paper is conducted in \cite{KamannBenchmarkingRobustnessSemantic2019}.
Kamann et al. \cite{KamannBenchmarkingRobustnessSemantic2019} provide a benchmark comparing several popular semantic segmentation models trained on CityScapes Dataset \cite{Cordts2016Cityscapes} and PASCAL VOC 2012 \cite{EveringhamPascalVisualObject2015}, then tested on corrupted versions of the same.
They proposed and verified that training models on noise can improve robustness to noise, but only reported ablation benchmarks across other types of corruption from \cite{hendrycks2018benchmarking}.

\input{_1_figures/dependence_structure.tex}

In an effort to continue bridging the gap between innovations made in robustness and semantic segmentation, we investigate the effects of corrupted images on a popular semantic segmentation model, DeeplabV3+ \cite{ChenDeepLabSemanticImage2018} with a Resnet50 \cite{HeDeepresiduallearning2016a} backbone.
Further, we provide evidence that improved robustness is a consequence of constructing pixel-wise representations with Implicit Background Estimation (IBE) from \cite{lehman2019implicit} and, even more so, with our proposed method to combine with the Sigmoid Cross Entropy objective (SCrIBE).
We validate our proposed method against the Baseline (Softmax Cross Entropy) and IBE (Softmax Cross Entropy with IBE) using the Imagenet-C Corruption Toolkit from \cite{hendrycks2018benchmarking} to corrupt the PASCAL VOC 2012 validation set \cite{EveringhamPascalVisualObject2015}.



%% file: _1_figures/dependence_structure.tex
\begin{figure}[t]
\input{_5_misc/dependence_structure.tex}%
\centering
\caption{Depictions of the dependency structures between components in $\vect{v}_{i,j}$ for each model.  The red node represents the background component, the red circle is the maximum foreground component, and red edges represents dependencies constructed with IBE. For IBE and SCrIBE, the background component is only dependent on the negative maximum foreground component, which results in a removal of $k-1$ dependencies.  In SCrIBE, the result is $k-1$ binary detectors that share the background representation.}
\label{fig:dependence}
\end{figure}
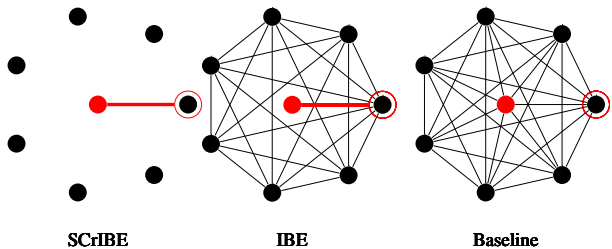

%% file: _5_misc/dependence_structure.tex
\scalebox{0.8}{
   \stargraph{7}{1.5}{SCrIBE}%
\fullyredstar{7}{1.5}{IBE}\quad%
  \fullygraph{7}{1.5}{Baseline}\quad%
}%

%% file: _4_sections/_2_background.tex
\section{Background}
\input{_1_figures/trainingflow.tex}

To better illustrate why IBE and SCrIBE benefit robustness, we must first review the structural properties of representation from the context of semantic segmentation. We will first analyze Softmax and provide insight on some properties that may result in susceptibility to corrupted inputs. Then, we will discuss IBE and outline why it results in the improved robustness observed in \cite{lehman2019implicit}. Finally, we will discuss why Sigmoid alone is a poor choice for optimizing a segmentation objective, but when combined with IBE it becomes superior in robustness to corruption.

\label{sec:representation}

\smpar{Semantic Segmentation} Let a semantic segmentation model be, $f(x):\R^{H\times W\times 3}\rightarrow \R^{H\times W \times k}$ and at each pixel location the output of $f(x)$ is $\vect{v}_{i,j}\in\R^k$, where $k$ is the number of classes.
$f(x)$ learns a representation for $\vect{v}_{i,j}$, which we define as $\K\in\R^k$. 
Also defined at each pixel is a categorical label, $\vect{y}_{i,j}$, as a one-hot vector.
Let softmax be, $\softmax_{SM}(\vect{x})$ and sigmoid be, $\softmax(\vect{x})$.

\subsection{Softmax}
Consider a model trained with Softmax Cross Entropy that achieves satisfactory task performance.
Beginning with the definition for the gradient of the Softmax Cross Entropy Loss for Semantic Segmentation in \eqref{eq:sm_grad}.  Let $\vect{x}^{(n)}$ be the scalar value of a vector indexed at $n$.

\begin{align}
\label{eq:sm_grad}
    \frac{\delta L}{\delta v_{i,j,n}} = \softmax_{SM}(\vect{v}_{i,j})^{(n)} - y_{i,j,n}.
\end{align}
During training, the updates follow the pattern of reinforcing True Positives (TP) by making that component larger and positive,  reinforcing True Negatives (TN) by making those components larger and negative, and punishing False Positives (FP) and False Negatives (FN) in the opposite way.
Though this behavior is desirable, Softmax heavily favors reinforcing TP and punishing FP because the updates for FN and TN depend on the relative magnitude of the components in $\vect{v}_{i,j}$ where $y_{i,j,n}=0$.
One insight is that those components associated with negative detection are neglected in optimization---more so as $k$ increases.
This is evident in Fig. \ref{fig:autocorrelation}, where the Baseline exhibits a chaotic structure in the autocorrelation of $\vect{v}_{i,j}^{13}$ because throughout training the FN and TN components were neglected leaving them close to the responses at initialization.
Another insight is that the structure of optimization depends only on the relative response in $\vect{v}_{i,j}$.
This structure is fully connected because detection depends on every component, $v_{i,j,n}$, as shown for the Baseline in Fig. \ref{fig:dependence}.

\input{_1_figures/autocorrelation.tex}

\input{_1_figures/explainedvariance.tex}

\subsection{Implicit Background Estimation}
When background is a class that must be learned---as with VOC2012---a problem arises with neglecting negative detection and requiring a fully connected dependency.
Background is the complement of the foreground classes.
In the case of VOC2012, the model must learn to represent potentially $k-1$ representations for background---one for each foreground class.
As was demonstrated in \cite{lehman2019implicit} and shown in Fig. \ref{fig:wholeflow}, by restricting detection of the background class to when all of the foreground components, $\vect{v}_{FG}\in\R^{k-1}$ are all in the negative orthant of $\R^{k-1}$, the surjections of Softmax are eliminated. 
Also, by inspecting the gradient update for IBE formulated as

\begin{align}
\label{eq:smibe_grad}
    \frac{\delta L}{\delta v_{i,j,n}} =  \begin{cases}
    -\frac{2 e^{v_{i,j,n}} \sum_{\text{FG}}e^{v_{i,j,m}} } {(\sum_{\text{FG}}e^{v_{i,j,m}})^2+1} \qquad n = \text{Background}&\\
    \softmax_{SM}(\vect{v}_{i,j})^{(n)} - y_{i,j,n} \qquad n \in \text{Foreground}&
    \end{cases}  .
\end{align}
We can observe that the TN and FN components for foreground classes are reinforced when the background is updated.
This is again evident in Fig. \ref{fig:autocorrelation}, where for both IBE and SCrIBE, there is orthogonality evident between component $13$ and all other components.
The outcome of this reinforcement will decorrelate TP detections allowing for the observed improvements in calibration and out-of-distribution detection in \cite{lehman2019implicit}.
However, as Softmax is still in use, the structure retains fully connected dependence for the foreground, and a binary dependence between the foreground classes and the shared background class as illustrated in Fig. \ref{fig:dependence}.

\subsection{Sigmoid}
To give some context on why Sigmoid is a poor choice for categorical classification consider the gradient update at $i,j$, which can be formulated as
    
\begin{align}
\label{eq:sg_grad}
    \frac{\delta L}{\delta v_{i,j,n}} = -y_{i,j,n}(1-\softmax(v_{i,j,n})).
\end{align}

\noindent
Notice that the update penalizes only for FN. Additionally, it allows for undesirable cases where $\vect{v}_{i,j}$ is a collection of large positive values---driving the $1-\softmax(\vect{v}_{i,j,n})$ term to 0---resulting in no gradient updates regardless the value of $y_{i,j,n}$. 
However, the key difference from Softmax, that we will utilize, is that Sigmoid "pins" component responses about 0, eliminating the relative dependence inherent with Softmax resulting in a structure of $k$ independent binary detectors.

%% file: _1_figures/trainingflow.tex
\begin{figure}[t]
\includegraphics[width=1.0\linewidth]{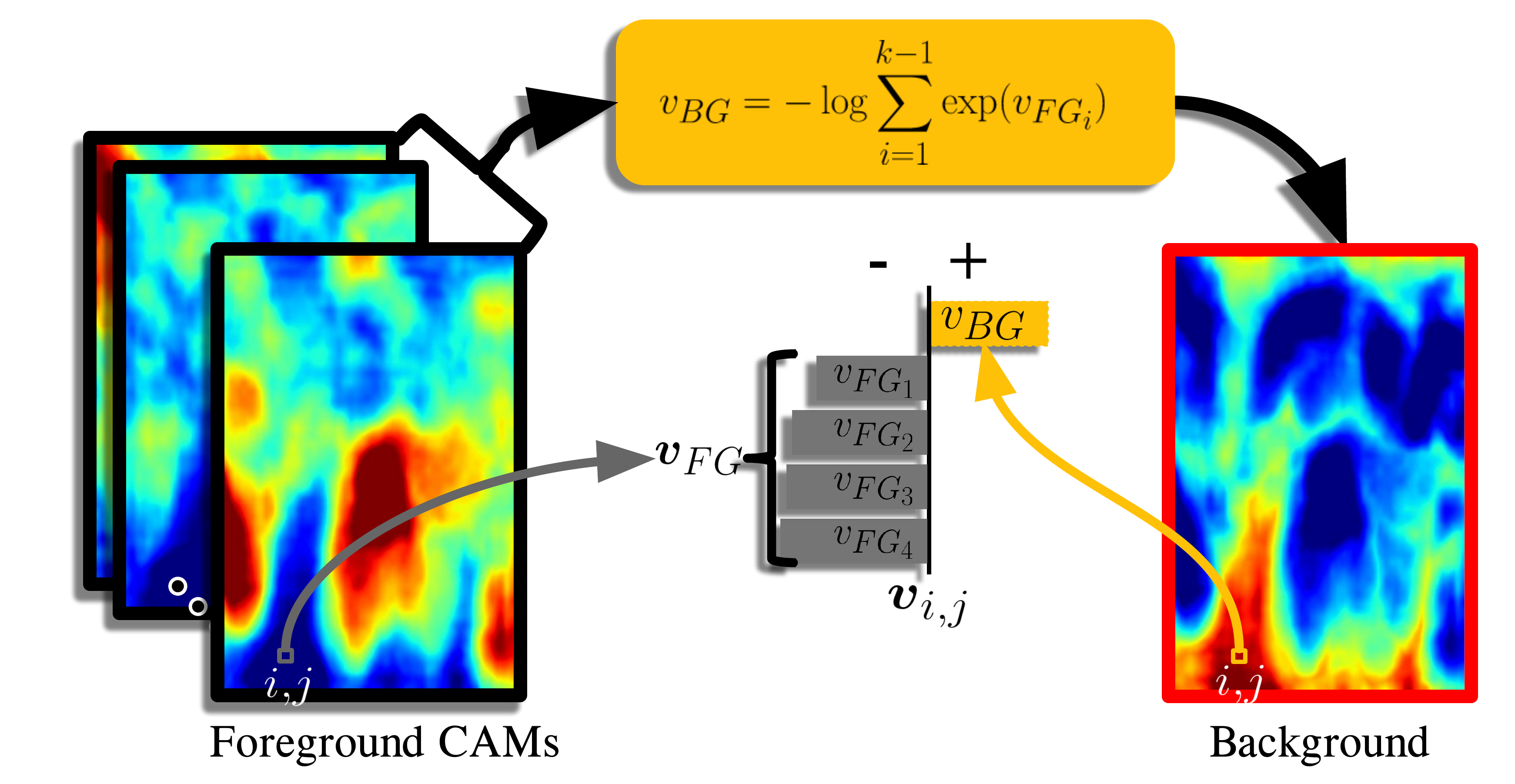}
\centering
\caption{The IBE module takes as an input the Class Activation Map (CAM) of the foreground, produces a background map and concatenates to the foreground CAMs to produce a complete prediction.}
\label{fig:wholeflow}
\end{figure}

%% file: _1_figures/autocorrelation.tex
\begin{figure}[htbp!]
\includegraphics[width=0.95\linewidth]{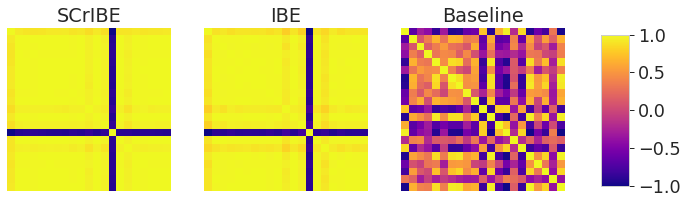}
\centering
\caption{Example auto-correlation matrices formed for each pixel, $\vect{v}_{i.j}^{(n)}$, where the prediction is $dog$ or $n=13$ of the validation set for each model trained with VOC2012.}

\label{fig:autocorrelation}
\end{figure}

%% file: _1_figures/explainedvariance.tex
\begin{figure}[htbp!]
\includegraphics[width=1.0\linewidth]{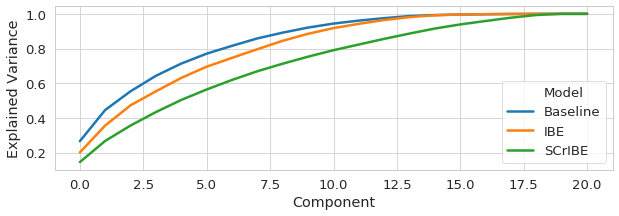}
\centering
\caption{Accumulated Explained Variance (EV) for all components in $\vect{v}_{i,j}$ by model.  Notice that both Baseline and IBE have lower dimensionality compared to SCrIBE due to the faster accumulation of EV.  SCrIBE retains dimensionality while  decorrelating predicted and non-predicted components.}

\label{fig:explainedvariance}
\end{figure}

%% file: _4_sections/_3_method.tex
\section{Sigmoid Cross Entropy with Implicit Background Estimation }
\label{sec:method}

When combining IBE with a Sigmoid Cross Entropy objective the gradient update becomes
\begin{align}
\label{eq:scribe_grad}
    \frac{\delta L}{\delta v_{i,j,n}} =  \begin{cases}
    -\frac{e^{v_{i,j,n}}}{1+\sum_{\text{FG}}e^{v_{i,j,m}}} \qquad\qquad n = \text{Background}\\
    -y_{i,j,n}(1-\softmax(v_{i,j,n}) \qquad n \in \text{Foreground}\\
    \end{cases}
\end{align} resulting in $k-1$ binary dependencies with only a single manifesting at a time, which is depicted in Fig. \ref{fig:dependence}.  
This arises from the background update enforcing the shared representation between each class and the Sigmoid operation acting on each component, $v_{i,j,n}$, independently.
Unlike the case of Sigmoid alone, training a collection of binary detectors with SCrIBE enforces an update to all foreground components when a background label is present.
The consequence of these updates is that the model must learn a rich representation for background to support foreground detection.
The resulting retention of dimensionality---notionally the "wiggle room"---should improves the robustness for corruptions that affect affine transformations in $\K$.

%% file: _4_sections/_4_experiment.tex
\section{Experimental Results}
\label{sec:experiment}
To validate the analysis in Section \ref{sec:representation} and test the hypothesis in Section \ref{sec:representation}, we evaluate all three versions of a single model.
We trained DeepLabV3+ with ResNet50 backbone on the unmodified VOC2012 training set augmented with the Semantic Boundaries Dataset (SBD) \cite{BharathICCV2011} to about 10k examples.
Each model was trained with images randomly scaled and cropped down to $224\times224$ pixels, a batch size of 30, "poly" scheduled learning rates starting at 0.01 for the backbone and 0.1 for the classifier, and weight decay of 5e-5. 
Note that the hardware used were $2\times$TITAN RTX with 24GB GPU memory each, but similar results are possible with a smaller batch size and lower learning rate.

\smpar{Inspection of Representation Structure}
To verify the observations made in analysis from \ref{sec:representation}, we compare the auto-correlation matrices and the accumulated explained variance for the data.  
Let $\vect{V}$ be the matrix formed by every pixel response for a $l$ number of inputs or  $\vect{V}^{(l)}=\vect{v}_{l,i,j}$.
The responses shown in Fig. \ref{fig:autocorrelation}, are normalized auto-correlation computed with $\vect{V}$.
Though only shown for a single class, these are consistent across all 21 classes.
It is clear that IBE is structuring the response away from the chaotic response shown for Baseline.
By decomposing the covariance matrix and inspecting the effective dimensionality of the learned representation, shown in Fig. \ref{fig:explainedvariance}, we can see the effect of using Sigmoid in the place of Softmax.
SCrIBE has an inherently higher-dimensional representation compared to Baseline and IBE, as is evident from the slower accumulation of Explained Variance.
In order to help support the notion that more "wiggle-room" results in resistance to the affine-in-$\K$ corruptions, we evaluate with a variety of corruptions.

\smpar{Robustness to Corrupted Input}
To test the robustness of each model, mean Intersection-over-Union (mIOU) metric was measured for the VOC 2012 validation set (1449 examples) that was corrupted using the first 15 corruptions in the ImageNet-C Corruption Toolkit at all 5 severity levels to create about 109k examples.
All models were tested on the same corrupted input simultaneously to remove effects caused by the variability in generating corruptions at runtime.
Additionally, the models were tested using the Multi-Scale Classification (MSC) method from \cite{ChenDeepLabSemanticImage2018} for comparison.
As summarized by Fig. \ref{fig:aggregatemiou} and Table \ref{tab:overallmiou} and  
\input{_2_tables/largeresults.tex}
\input{_1_figures/aggregatemiou.tex}
\input{_1_figures/corruptionsummary.tex}
detailed in Table \ref{tab:CorruptFull}, SCrIBE is clearly superior to the Baseline across almost all corruptions and severity levels, while all models have very similar performance with no corruption.
However, SCrIBE does not improve across all corruptions compared to IBE.
As suggested earlier, the additional "wiggle-room" gained by the Sigmoid objective will only help with affine-in-$\K$ corruptions.
MSC improves the performance for all models, as shown in Fig. \ref{fig:aggregatemiou}.

\smpar{Qualitative Results}
We provide visualizations comparing the effects of 3 corruptions between Bseline and SCrIBE for qualitiative evaluation in Fig. \ref{fig:corruptionsummary}.
In general for semantic segmentation, robustness manifest as retention of predictions under corrupted conditions.
For all models, it is more often the case that a misclassification is to background and not some other foreground class.
However, as there are some cases where a foreground class is the resulting misclassification, we have observed that these exchanges follow the relative label frequency of the training set.

%% file: _2_tables/largeresults.tex
\begin{table*}[htbp!]
		 \setlength{\tabcolsep}{4pt}
		 \centering
		 \begin{tabularx}{\textwidth}{
		  *{1}{>{\hsize=0.03\hsize}Y} 
		  *{1}{>{\hsize=0.10\hsize}Y}
		 |*{3}{>{\hsize=0.06\hsize}Y}
		 |*{4}{>{\hsize=0.06\hsize}Y}
		 |*{3}{>{\hsize=0.06\hsize}Y}
		 |*{2}{>{\hsize=0.06\hsize}Y}
		 |*{3}{>{\hsize=0.06\hsize}Y} }
		 \Xhline{8\arrayrulewidth} 
		 \multicolumn{2}{c}{} & \multicolumn{3}{c}{Noise} & \multicolumn{4}{c}{Blur}  & \multicolumn{3}{c}{Weather} & \multicolumn{2}{c}{Lighting} & \multicolumn{3}{c}{Spatial}\\ 
		 \cline{3-11}
		 \Xhline{8\arrayrulewidth} 
		 {Sv.} & {Model} & {Gaus.} & {Sh.} & {Imp.} & {Dfc.} & {Gls.} & {Mtn.} & {Zm.} & {Sno.} & {Frs.} & {Fog} & {Bri.} & {Cnt.} & {Ela.} & {Pix.} & {JPEG}\\ \hline
		 \parbox[t]{50mm}{\multirow{3}{*}{1}}
& Baseline &  55.5 &  56.0 &  50.0 &  52.5 &  49.5 &  56.0 &  44.5 &  51.0 &  60.0 &  63.0 &  69.5 &  66.5 &  48.5 &  62.0 &  62.5 \\
& IBE &  59.0 &  59.0 &  50.5 &  57.5 & \bf 53.0 &  59.5 &\bf  49.0 &  52.0 &  61.0 &\bf  65.5 &\bf  71.0 &\bf  68.5 &  49.5 &  60.5 &  63.0 \\
& SCrIBE & \bf 60.5 & \bf 61.5 &\bf  54.5 &\bf  61.5 &  48.5 &\bf  61.0 &\bf  49.0 &\bf  53.5 &\bf  61.5 &  64.5 &  70.0 &\bf  68.5 &\bf  50.5 &\bf  63.5 &\bf  63.5 \\
		 \Xhline{3\arrayrulewidth}
		 \parbox[t]{50mm}{\multirow{3}{*}{2}}
& Baseline &  43.0 &  41.5 &  37.0 &  41.0 &  33.5 &  41.5 &  35.5 &  32.5 &  44.0 &  60.0 &  69.0 &  63.0 &  25.5 &  59.0 &  59.0 \\
& IBE &  48.5 &  47.0 &  40.5 &  49.5 & \bf 39.0 &  47.5 &\bf  40.5 &  33.5 &\bf  45.5 & \bf 62.0 &\bf  70.0 &\bf  66.5 &  26.5 &  58.0 &  59.0 \\
& SCrIBE & \bf 51.0 &\bf  50.0 &\bf  44.0 &\bf  54.5 &  31.5 &\bf  50.5 &  40.0 &\bf  34.5 &\bf  45.5 &\bf  62.0 &  69.0 &  66.0 &\bf  28.5 &\bf  62.0 &\bf  61.0 \\
		 \Xhline{3\arrayrulewidth}
		 \parbox[t]{50mm}{\multirow{3}{*}{3}}
& Baseline &  26.0 &  26.0 &  26.0 &  22.5 &  12.0 &  24.5 &  29.5 &  36.0 &  33.5 &  54.0 &  67.5 &  55.0 &  51.0 &  38.0 &  56.5 \\
& IBE &  30.5 &  29.5 &  29.0 &  33.5 &\bf  14.5 &  30.5 &\bf  36.0 &\bf  39.0 &\bf  35.0 &\bf  57.0 &\bf  68.0 &  60.0 & \bf 53.0 &  38.5 &  56.5 \\
& SCrIBE &\bf  34.5 &\bf  34.5 &\bf  33.5 &\bf  39.0 &  11.0 &\bf  34.0 &  35.5 &\bf  39.0 &  34.5 &\bf  57.0 &\bf  68.0 &\bf  61.5 &  50.5 &\bf  46.5 &\bf  58.0 \\
		 \Xhline{3\arrayrulewidth}
		 \parbox[t]{50mm}{\multirow{3}{*}{4}}
& Baseline &  10.5 &   9.0 &   8.5 &  12.5 &   8.0 &  12.5 &  24.0 &  27.0 &  31.5 &  49.5 &  64.5 &  34.5 &  35.0 &  19.5 &  45.0 \\
& IBE &  12.5 &  12.0 &  10.5 &  21.0 &\bf  10.5 &  16.0 &\bf  29.0 &\bf  30.5 &\bf  33.5 &  52.0 &  65.5 &  41.5 &\bf  36.5 &  21.0 &  46.5 \\
& SCrIBE &\bf  17.5 &\bf  14.5 &\bf  15.0 &\bf  24.5 &   8.0 &\bf  18.5 &\bf  29.0 &  30.0 &  32.5 &\bf  52.5 &\bf  66.0 &\bf  45.0 &  34.5 &\bf  28.5 &\bf  50.0 \\
		 \Xhline{3\arrayrulewidth}
		 \parbox[t]{50mm}{\multirow{3}{*}{5}}
& Baseline &   3.5 &   4.5 &   3.5 &   7.5 &   5.0 &   8.5 &  19.0 &  24.3 &  25.5 &  34.0 &  60.5 &  14.0 &  13.5 &  13.5 &  29.0 \\
& IBE &   5.0 &   6.5 &   4.5 &  12.5 &\bf   7.5 &  11.0 &\bf  24.5 &\bf  26.0 & \bf 28.0 &  39.0 &\bf  62.5 &  20.0 &\bf  15.5 &  14.5 &  32.0 \\
& SCrIBE &\bf   5.5 &\bf   7.5 &\bf   5.5 &\bf  14.5 &   6.0 &\bf  13.0 &  23.0 &  24.3 &  26.5 &\bf  40.0 &\bf  62.5 &\bf  23.0 &\bf  15.5 &\bf  20.5 &\bf  36.7 \\ 
		 \Xhline{8\arrayrulewidth}
		 \parbox[t]{50mm}{\multirow{3}{*}{\rotatebox{90}{Mean}}}
& Baseline &  37.4 &  37.0 &  35.4 &  36.7 &  21.9 &  29.0 &  30.6 &  32.6 &  39.7 &  52.7 &  66.9 &  46.5 &  35.2 &  39.8 &  51.4 \\
& IBE &  40.2 &  40.0 &  36.7 &  43.4 &\bf  24.8 &  33.3 &\bf  35.8 &\bf  34.7 &  41.0 &\bf  55.6 &\bf  68.2 &  51.3 &  36.4 &  40.5 &  52.3 \\
& SCrIBE &\bf  43.5 &\bf  43.3 &\bf  41.2 &\bf  47.3 &  21.3 &\bf  35.5 &  35.5 &  34.6 &\bf  41.4 &  55.3 &  67.9 &\bf  52.9 &\bf  36.6 &\bf  46.3 &\bf  55.8 \\
		 \Xhline{8\arrayrulewidth}
\end{tabularx}
		 \caption{mIOU scores for corrupted VOC 2012 validation set. SCrIBE is clearly superior to the Baseline on almost every corruption type and level.  Compared to IBE, SCrIBE offers improvements to Noise, and Spatial robustness, however, improvement is mixed otherwise.}
		 \label{tab:CorruptFull}
		 \end{table*}

%% file: _1_figures/aggregatemiou.tex
\begin{figure}[t]
\includegraphics[width=\linewidth]{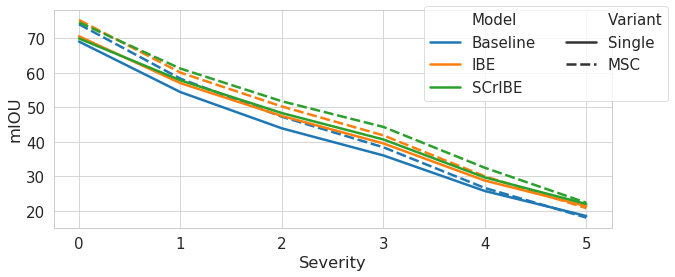}
\centering
\caption{This plot aggregates mIOU across each corruption to show the effects of corruption severity.  Severity 0 indicates uncorrupted inputs.  With Multi-Scale Classification (MSC) both models have similar performance at Severity 0, but only SCrIBE continues to benefit from MSC.  The Baseline actually performs worse when MSC is used for corrupted inputs.}
\label{fig:aggregatemiou}
\end{figure}

%% file: _1_figures/corruptionsummary.tex
\begin{figure}[ht!]
\includegraphics[trim={0 27.7cm 0 0},clip, width=0.96\linewidth]{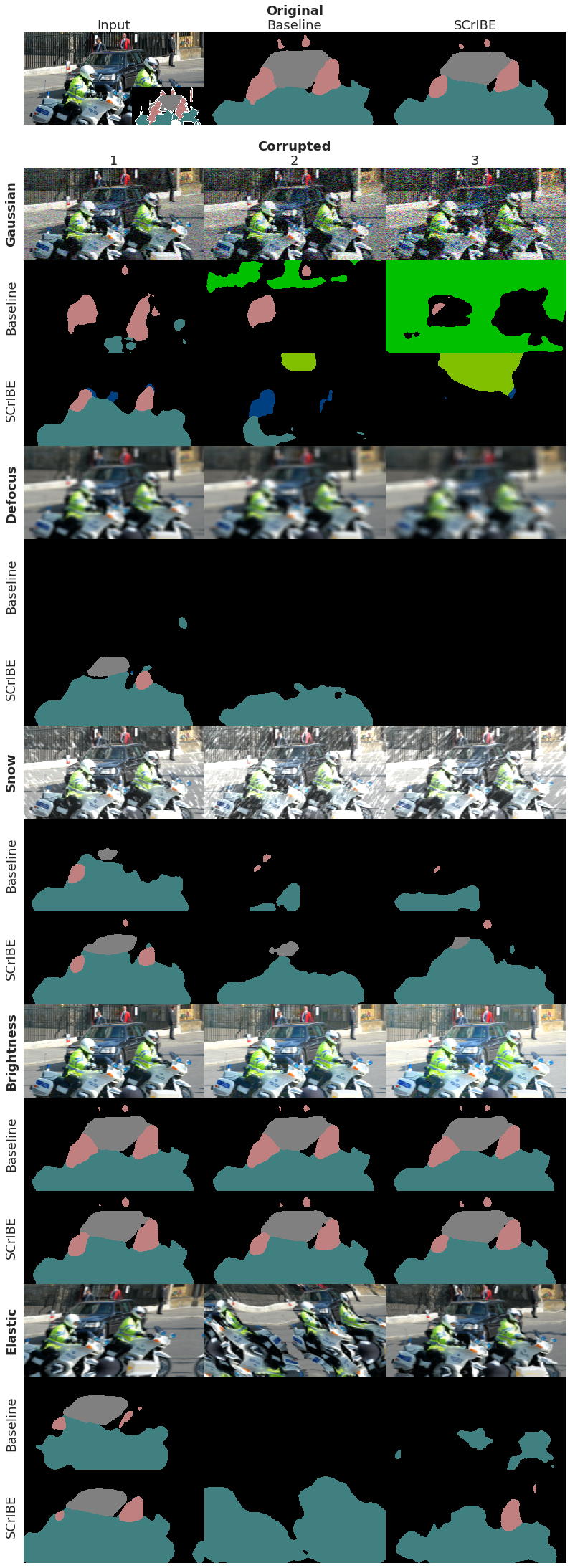}
\centering
\caption{The (top) row, labeled Original, shows a comparison between Baseline and SCrIBE for the input.  The proceeding rows, labeled Corrupted, shows a comparison across the first 3 severity levels for the 3 corruptions. Though severities 4 and 5 are not shown, it is clear by the results in Table \ref{tab:CorruptFull}, the predictions are generally entirely incorrect.}
\label{fig:corruptionsummary}
\end{figure}

%% file: _4_sections/_5_conclusion.tex
\section{Conclusion}
\label{sec:conclusion}
In this paper, we provided analytical and empirical evidence about the underlying structures of representation for semantic segmentation.
We determined that the decorrelated components produced by IBE result in improved robustness.
From the analysis, we hypothesised about corruptions that act as affine transformations in the representation space $\K$.
We then showed that the effects of affine-in-$\K$ corruptions can be further reduced by retaining dimensionality of the representation through applying SCrIBE.
Though these results are promising for improving the robustness of semantic segmentation models, the evidence suggests that the properties of representation structure are largely unknown and likely untapped.
Namely, though the addition of IBE and SCrIBE did improve robustness to corruptions and evidence was presented to associate observable properties with the improvement, the question of direct causality is still unanswered. 
\input{_2_tables/baseline_miou.tex}

%% file: _2_tables/baseline_miou.tex
\begin{table}[t]
\centering
\begin{tabular}{r|cccc}
\toprule
\textbf{Model}&\textbf{val}&\textbf{val+MSC}&\textbf{cor}&\textbf{cor+MSC} \\ \toprule
Baseline  & 69.1 & 74.1 & 35.5 & 37.5\\ 
IBE & \bf 70.6 & \bf 75.3& 38.6 & 40.3\\ 
SCrIBE & 69.9 & 74.6& \textbf{39.5} & \textbf{42.1}\\ 
\hline
\end{tabular}
\caption{Results in terms of mIOU on PASCAL VOC 2012 validation set using DeepLabv3+ with ResNet-50 backbone for Baseline and our SCrIBE variant.  Multi-Scale Classification (MSC) is also used to improve performance. The results are aggregated across 15 corruptions at 5 severity levels are provided for Baseline, IBE, SCrIBE both with and without MSC.  }
\label{tab:overallmiou}
\end{table}

%% file: main.bbl
\begin{thebibliography}{10}

\bibitem{Zion2019}
``Machine learning market by service (professional services, and managed
  services): Global industry perspective, comprehensive analysis, and forecast,
  2017-2024',''
\newblock Tech. {R}ep., {Zion Market Research}, 2019.

\bibitem{hatcher2018survey}
W.~G. Hatcher and W. Yu,
\newblock ``A survey of deep learning: Platforms, applications and emerging
  research trends,''
\newblock {\em IEEE Access}, vol. 6, pp. 24411--24432, 2018.

\bibitem{pytorch2019}
A. Paszke et~al.,
\newblock ``Pytorch: An imperative style, high-performance deep learning
  library,''
\newblock in {\em Advances in Neural Information Processing Systems 32}, pp.
  8024--8035. Curran Associates, Inc., 2019.

\bibitem{tensorflow2015-whitepaper}
M. Abadi et~al.,
\newblock ``{TensorFlow}: Large-scale machine learning on heterogeneous
  systems,'' 2015,
\newblock Software available from tensorflow.org.

\bibitem{ntsbHWY18MH010}
N.~T.~S. Board,
\newblock ``Collision between vehicle controlled by developmental automated
  driving system and pedestrian,''
\newblock Tech. {R}ep., {NTSB}, 2019.

\bibitem{KamannBenchmarkingRobustnessSemantic2019}
C. Kamann and C. Rother,
\newblock ``Benchmarking the {{Robustness}} of {{Semantic Segmentation
  Models}},''
\newblock {\em arXiv:1908.05005 [cs, eess]}, Aug. 2019.

\bibitem{hendrycks17baseline}
D. Hendrycks and K. Gimpel,
\newblock ``A baseline for detecting misclassified and out-of-distribution
  examples in neural networks,''
\newblock {\em Proceedings of International Conference on Learning
  Representations}, 2017.

\bibitem{liang2018enhancing}
S. Liang, Y. Li, and R. Srikant,
\newblock ``Enhancing the reliability of out-of-distribution image detection in
  neural networks,''
\newblock in {\em International Conference on Learning Representations}, 2018.

\bibitem{lee2018simple}
K. Lee, K. Lee, H. Lee, and J. Shin,
\newblock ``A simple unified framework for detecting out-of-distribution
  samples and adversarial attacks,''
\newblock in {\em Advances in Neural Information Processing Systems}, 2018, pp.
  7167--7177.

\bibitem{hendrycks2018oe}
D. Hendrycks et~al.,
\newblock ``Deep {{Anomaly Detection}} with {{Outlier Exposure}},''
\newblock {\em arXiv preprint}, 2018.

\bibitem{guo2017calibration}
C. Guo et~al.,
\newblock ``On calibration of modern neural networks,''
\newblock in {\em Proceedings of the 34th International Conference on Machine
  Learning-Volume 70}. {JMLR. org}, 2017, pp. 1321--1330.

\bibitem{lee2018training}
K. Lee et~al.,
\newblock ``Training confidence-calibrated classifiers for detecting
  out-of-distribution samples,''
\newblock in {\em International Conference on Learning Representations}, 2018.

\bibitem{hendrycks2018benchmarking}
D. Hendrycks and T. Dietterich,
\newblock ``Benchmarking neural network robustness to common corruptions and
  perturbations,''
\newblock in {\em International Conference on Learning Representations}, 2019.

\bibitem{Carlinievaluatingrobustnessneural2017}
N. Carlini and D. Wagner,
\newblock ``Towards evaluating the robustness of neural networks,''
\newblock in {\em Security and {{Privacy}} ({{SP}}), 2017 {{IEEE Symposium}}
  On}. 2017, pp. 39--57, {IEEE}.

\bibitem{papernot2016limitations}
N. Papernot et~al.,
\newblock ``The limitations of deep learning in adversarial settings,''
\newblock in {\em 2016 IEEE European symposium on security and privacy
  (EuroS\&P)}. IEEE, 2016, pp. 372--387.

\bibitem{GoodfellowExplainingharnessingadversarial2014}
I.~J. Goodfellow et~al.,
\newblock ``Explaining and harnessing adversarial examples,''
\newblock {\em arXiv preprint arXiv:1412.6572}, 2014.

\bibitem{Temel2018_CUREOR}
D. Temel, J. Lee, and G. AlRegib,
\newblock ``{CURE-OR: Challenging Unreal and Real Environments for Object
  Recognition},''
\newblock in {\em IEEE International Conference on Machine Learning and
  Applications (ICMLA)}, 2018.

\bibitem{temel2019traffic}
D. Temel, M. Chen, and G. AlRegib,
\newblock ``Traffic sign detection under challenging conditions: {{A}} deeper
  look into performance variations and spectral characteristics,''
\newblock {\em IEEE Transactions on Intelligent Transportation Systems}, pp.
  1--11, 2019.

\bibitem{GeirhosGeneralisationhumansdeepa}
R. Geirhos et~al.,
\newblock ``Generalisation in humans and deep neural networks,''
\newblock p.~13.

\bibitem{Temel2017_NIPSW}
D. Temel, G. Kwon, M. Prabhushankar, and G. AlRegib,
\newblock ``{CURE-TSR: Challenging unreal and real environments for traffic
  sign recognition},''
\newblock in {\em Neural Information Processing Systems (NeurIPS) Workshop on
  Machine Learning for Intelligent Transportation Systems}, 2017.

\bibitem{Temel2018_SPM}
D. Temel and G. AlRegib,
\newblock ``Traffic signs in the wild: Highlights from the ieee video and image
  processing cup 2017 student competition [sp competitions],''
\newblock {\em IEEE Sig. Proc. Mag.}, vol. 35, no. 2, pp. 154--161, 2018.

\bibitem{Temel2019_ICIP}
D. {Temel}, J. {Lee}, and G. {AlRegib},
\newblock ``Object recognition under multifarious conditions: A reliability
  analysis and a feature similarity-based performance estimation,''
\newblock in {\em 2019 IEEE International Conference on Image Processing
  (ICIP)}, Sep. 2019, pp. 3033--3037.

\bibitem{EveringhamPascalVisualObject2015}
M. Everingham et~al.,
\newblock ``The {{Pascal Visual Object Classes Challenge}}: {{A
  Retrospective}},''
\newblock {\em International Journal of Computer Vision}, vol. 111, no. 1, pp.
  98--136, Jan. 2015.

\bibitem{LinMicrosoftCOCOCommon2015}
T.-Y. Lin et~al.,
\newblock ``Microsoft {{COCO}}: {{Common Objects}} in {{Context}},''
\newblock {\em arXiv:1405.0312 [cs]}, Feb. 2015.

\bibitem{Cordts2016Cityscapes}
M. Cordts et~al.,
\newblock ``The cityscapes dataset for semantic urban scene understanding,''
\newblock in {\em Proc. of the IEEE Conference on Computer Vision and Pattern
  Recognition (CVPR)}, 2016.

\bibitem{ZhouSceneparsingade20k2017}
B. Zhou et~al.,
\newblock ``Scene parsing through ade20k dataset,''
\newblock in {\em Proc. {{CVPR}}}, 2017.

\bibitem{SiamRTSegRealTimeSemantic2018}
M. Siam et~al.,
\newblock ``{{RTSeg}}: {{Real}}-{{Time Semantic Segmentation Comparative
  Study}},''
\newblock in {\em 2018 25th {{IEEE International Conference}} on {{Image
  Processing}} ({{ICIP}})}, {Athens}, Oct. 2018, pp. 1603--1607, {IEEE}.

\bibitem{MehtaESPNetEfficientSpatial2018}
S. Mehta et~al.,
\newblock ``{{ESPNet}}: {{Efficient Spatial Pyramid}} of {{Dilated
  Convolutions}} for {{Semantic Segmentation}},''
\newblock in {\em Computer {{Vision}} \textendash{} {{ECCV}} 2018}, vol. 11214,
  pp. 561--580. {Springer International Publishing}, {Cham}, 2018.

\bibitem{HowardMobileNetsEfficientConvolutional2017}
A.~G. Howard et~al.,
\newblock ``{{MobileNets}}: {{Efficient Convolutional Neural Networks}} for
  {{Mobile Vision Applications}},''
\newblock {\em arXiv:1704.04861 [cs]}, Apr. 2017.

\bibitem{PaszkeENetDeepNeural2016}
A. Paszke et~al.,
\newblock ``{{ENet}}: {{A Deep Neural Network Architecture}} for
  {{Real}}-{{Time Semantic Segmentation}},''
\newblock {\em arXiv:1606.02147 [cs]}, June 2016.

\bibitem{ZhouAutomatedEvaluationSemantic2019}
W. Zhou et~al.,
\newblock ``Automated {{Evaluation}} of {{Semantic Segmentation Robustness}}
  for {{Autonomous Driving}},''
\newblock {\em IEEE Transactions on Intelligent Transportation Systems}, pp.
  1--13, 2019.

\bibitem{lehman2019implicit}
C. Lehman, D. Temel, and G. AlRegib,
\newblock ``Implicit background estimation for semantic segmentation,''
\newblock in {\em 2019 {{IEEE}} International Conference on Image Processing
  ({{ICIP}})}. {IEEE}, 2019, pp. 1935--1939.

\bibitem{ChenDeepLabSemanticImage2018}
L.-C. Chen et~al.,
\newblock ``{{DeepLab}}: {{Semantic Image Segmentation}} with {{Deep
  Convolutional Nets}}, {{Atrous Convolution}}, and {{Fully Connected CRFs}},''
\newblock {\em IEEE Transactions on Pattern Analysis and Machine Intelligence},
  vol. 40, no. 4, pp. 834--848, Apr. 2018.

\bibitem{HeDeepresiduallearning2016a}
K. He, X. Zhang, S. Ren, and J. Sun,
\newblock ``Deep residual learning for image recognition,''
\newblock in {\em Proceedings of the {{IEEE}} Conference on Computer Vision and
  Pattern Recognition}, 2016, pp. 770--778.

\bibitem{BharathICCV2011}
B. Hariharan, P. Arbelaez, L. Bourdev, S. Maji, and J. Malik,
\newblock ``Semantic contours from inverse detectors,''
\newblock in {\em International Conference on Computer Vision (ICCV)}, 2011.

\end{thebibliography}
